\documentclass[amsmath,amssymb,amsfonts,pra,aps,floatfix,11pt,tightenlines,raggedbottom,superscriptaddress]{revtex4-2}
\usepackage{hyperref}
\usepackage{multirow}
\usepackage{booktabs}
\usepackage{graphicx}
\usepackage[svgnames,x11names,table]{xcolor}
\usepackage{hyperref}
\usepackage[protrusion=true,expansion=true]{microtype} 
\usepackage{placeins}

\hypersetup{
    colorlinks=true,
    urlcolor=MediumBlue,
    linkcolor=MediumBlue,
    citecolor=MediumBlue,
    filecolor=MediumBlue,
    pdftitle={Visual Odometry with Neuromorphic Resonator Networks},
    pdfsubject={cs.RO},
    pdfauthor={Alpha Renner},
    pdfkeywords={Visual Odometry, Hyperdimensional Computing, Vector Symbolic Architecture (VSA), Event-Based Vision, Neuromorphic Hardware, Spiking Neural Networks (SNN), Loihi, Neuromorphic Algorithms, Robotics, Visual SLAM},
}

\begin{document}
\title{Visual Odometry with Neuromorphic Resonator Networks}
\author{Alpha Renner} \email{alpren@ini.uzh.ch}
 \affiliation{Institute of Neuroinformatics, University of Zurich and ETH Zurich, Switzerland}
 \affiliation{Forschungszentrum Jülich, Germany}
\author{Lazar Supic}\email{lsupic@gmail.com}
\affiliation{Accenture Labs, San Francisco, CA, USA}
\author{Andreea Danielescu}
\affiliation{Accenture Labs, San Francisco, CA, USA}
\author{Giacomo Indiveri}
 \affiliation{Institute of Neuroinformatics, University of Zurich and ETH Zurich, Zürich, Switzerland}
\author{E. Paxon Frady}
\affiliation{Intel Neuromorphic Computing Lab, Intel Labs, Santa Clara, CA, USA}
\author{Friedrich T. Sommer}\email{fsommer@berkeley.edu}
\affiliation{Redwood Center for Theoretical Neuroscience, UC Berkeley, CA, USA}
\affiliation{Intel Neuromorphic Computing Lab, Intel Labs, Santa Clara, CA, USA}
\author{Yulia Sandamirskaya}\email{yulia.sandamirskaya@zhaw.ch}
\affiliation{Intel Labs, Zürich, Switzerland}
\affiliation{ZHAW Zurich University of Applied Sciences, W\"adenswil, Switzerland}

\begin{abstract}
Visual Odometry (VO) is a method to estimate self-motion of a mobile robot using visual sensors. Unlike odometry based on integrating differential measurements that can accumulate errors, such as inertial sensors or wheel encoders, visual odometry is not compromised by drift. However, image-based VO is computationally demanding, limiting its application in use cases with low-latency, low-memory, and low-energy requirements.
Neuromorphic hardware offers low-power solutions to many vision and AI problems, but designing such solutions is complicated and often has to be assembled from scratch. 
Here we propose to use Vector Symbolic Architecture (VSA) as an abstraction layer to design algorithms compatible with neuromorphic hardware. Building from a VSA model for scene analysis, described in our companion paper, we present a modular neuromorphic algorithm that achieves state-of-the-art performance on two-dimensional VO tasks.
Specifically, the proposed algorithm stores and updates a working memory of the presented visual environment. Based on this working memory, a resonator network estimates the changing location and orientation of the camera. 
We experimentally validate the neuromorphic VSA-based approach to VO with two benchmarks: one based on an event camera dataset and the other in a dynamic scene with a robotic task.
\end{abstract}
\flushbottom
\maketitle
\thispagestyle{empty}

Animals as small as bees, with less than one million neurons, show an extraordinary ability to navigate complex environments using visual information~\cite{Srinivasan2010bees}. These animals use visual signals to estimate their motion and to keep track of their position relative to important locations. A comparable computation performed by machines is called  Visual Odometry (VO).  The biological solution to VO remains unmatched in compactness and energy efficiency compared to today's best technical solutions found in robotics \cite{nister2004visual,scaramuzza2011visual}. 
Improving the energy efficiency of VO is an open challenge that can enable novel applications, such as small autonomous drones~\cite{palossi_2019}, planetary rovers~\cite{moravec1980obstacle, lacroix1999rover, corke2004omnidirectional}, or light-weight augmented reality (AR) glasses.
Current VO algorithms are mostly implemented using image-based sensors (cameras) and frame-based, synchronous computing, either with an on-device CPU or GPU or in the cloud. In both cases, image processing for VO is a computationally demanding operation. It requires computing spatial correlation between subsequent frames, directly between pixels, or between extracted features that can be tracked reliably. Improving the efficiency of VO can enable advanced navigation capabilities on small autonomous robotic devices with a limited power budget. The working principles of biological visual self-motion estimation are a promising avenue to build novel, more efficient neural-inspired VO processing systems.

The field of neuromorphic engineering pursues the goal of building computing hardware that emulates biological signal processing~\cite{mead_1990,chicca_2014}. The main insight in this field is that the efficiency of brains exploits a close match between algorithms and the structure of the biological computing hardware. Neuromorphic processors typically consist of many parallel digital or mixed-signal computing elements that reproduce the dynamics of biological spiking neurons and the synapses connecting them. These hardware architectures use asynchronous, event-driven communication between neurons and fine-grained parallelism with local memory. This yields power-efficient implementations of neural network-based algorithms, in some cases showing orders of magnitude advantages in power consumption and time to solution~\cite{davies2021advancing, Indiveri_Liu_2015}.

The main challenge in exploiting the full potential of neuromorphic hardware is developing a computing framework. Such a framework needs to provide an abstraction layer to link a desired computation to a neural or neuromorphic implementation, i.e., for programming the many features of neurons and synapses on neuromorphic hardware and connecting neurons in networks to solve a given computational task. Based on connectionist models for symbolic reasoning~\cite{PlateAnalogy2000,gayler_1998,KanervaOrdered1996} originating in cognitive science and theoretical neuroscience, several such frameworks have been proposed \cite{eliasmith2003neural, sandamirskaya2013dynamic, KleykoComputingParadigm2021,liang2019neural}.
The abstraction is achieved by designating populations of neurons to represent behavioral variables or symbols and defining (neural or synaptic) operations for manipulating these variables.

A family of such frameworks~\cite{eliasmith2003neural, KleykoComputingParadigm2021} are known as Vector Symbolic Architectures (VSA)~\cite{GaylerJackendoff2003} or hyperdimensional computing \cite{kanerva_2009}. 
In these frameworks, variables, symbols, and operations are represented by high-dimensional vectors. Vectors can be composed of binary, integer, real, or complex numbers, depending on the VSA type used. The high dimensionality of vectors in VSA means that random vectors representing different variables are almost orthogonal. This enables the encoding of composite data structures into single vectors. A neuromorphic algorithm is then defined by vector operations selectively decoding and manipulating variables in such data structures. 

In this work, we use a VSA called Fourier Holographic Reduced Representation~\cite{plate1995holographic} that computes with complex vectors. Based on a spike-timing code for complex numbers \citep{frady_2019tpam}, we show in the accompanying paper~\cite{renner2022T} how complex vector computations can be implemented in populations of spiking neurons on neuromorphic hardware. Specifically, the phase of a complex number with unit amplitude, a so-called phasor~\cite{noest_1988}, is encoded by the timing of a neuron's spike.
In this work, we present and benchmark a proof-of-principle robotic application designed with such a VSA framework.

Computationally, visual odometry amounts to estimating geometrical transformations between pairs of images obtained with a moving camera. This computation can be used in other tasks requiring estimating the transformation between two or more views, e.g., for scene reconstruction or video compression. Recent developments using VSA representations, fractional power encoding \cite{plate1995holographic,frady_2019precession, frady2021VFA, komer_2019} and resonator networks \citep{frady_2020_resonator} allow us to estimate such transformations efficiently and with low latency. 
Our results on odometry suggest that the proposed phasor VSA framework with its recent extensions \cite{frady2021VFA} provides the necessary robustness and flexibility for solving computationally challenging robotics applications and may, therefore, be a good candidate for a neuromorphic realization of the VSA computing framework.

\begin{figure}[ht]
  \centering
  \includegraphics[width=0.8\linewidth]{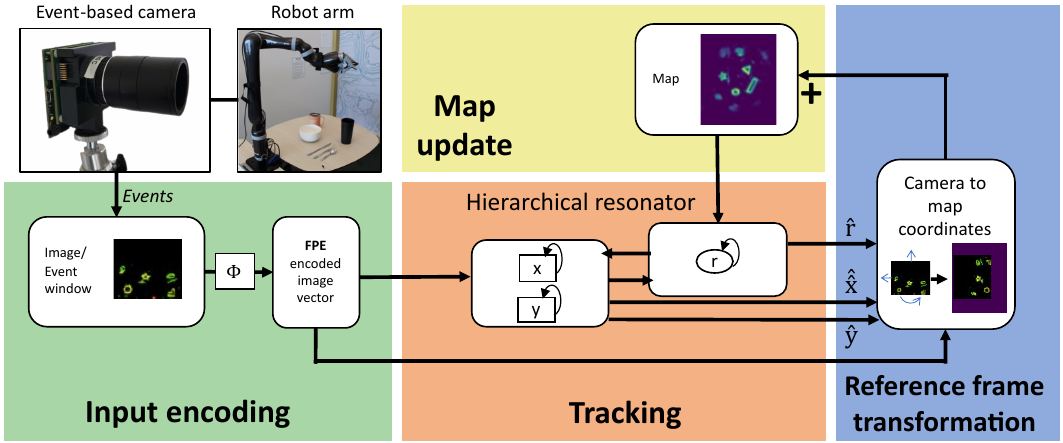}
  \caption{\textit{Neuromorphic Event-Based Visual Odometry with the Hierarchical Resonator Network.} Events from an event-based camera are collected and encoded using fractional power encoding (FPE) into a VSA vector. The generative model assumes the input is generated from a rotated and translated map. This model is inverted by the hierarchical resonator using two interacting partitions with different frames of reference. The translation ($\hat{x},\hat{y}$) and rotation ($\hat{r}$) (and optionally, scale) estimates are then used to transform the input into map coordinates and to update the map.}
  \label{fig:overview}
\end{figure}

\section*{Results}
\subsection*{VSA-based approach to visual odometry} \label{sec:HRN-VO}
The VSA-based approach to VO that we develop here leverages a neuromorphic method for scene analysis based on a generative model described in our accompanying paper~\cite{renner2022T}. 
A generative model of a scene holds knowledge of object shapes and their transforms, which may generate different images.
The analysis of a given image requires inference in the generative model, involving a computationally expensive search over all shapes and transforms.  

The method in~\cite{renner2022T} uses a formulation of the generative model in which an image is represented as a sum of products of VSA vectors. The terms in the sum correspond to different objects in the image. The vectors in the products describe different factors of variation of each individual object, such as shape, translation, and rotation. 

In particular, inference constitutes vector factorization, which can be performed efficiently and in parallel on neuromorphic hardware~\cite{renner2022T} by the so-called resonator network \cite{frady_2020_resonator,kent_2020resonator}.  Specifically, resonator networks can perform pattern recognition invariant with respect to a certain image transform, such as translation or rotation. To enable this, the image is encoded so that image transforms can be expressed as vector binding. Image translation requires encoding into a Cartesian reference frame by a transform that fulfills the Fourier convolution theorem. The convolution theorem then guarantees that each translation in the image domain can be computed on the image representations by component-wise multiplication with a vector, that is, by vector binding. We do this in~\cite{renner2022T} by using fractional power encoding \cite{frady_2019precession,komer_2019,frady2021VFA} to encode the input image. This linear transform into the complex domain is similar to the conventional Fourier transform but with randomized rather than regularly spaced frequencies.

Standard resonator networks cannot perform inference for generative models of images that combine factors of variation that do not commute, such as rotation and translation transforms. For inference in such generative models of images, \cite{renner2022T} proposes a resonator network with a novel partitioned architecture, the hierarchical resonator network (HRN). 
The hierarchical resonator network consists of two partitions that work in different reference frames: Cartesian and log-polar. The two partitions communicate via matrix transforms that convert from one reference frame to the other. 

The log-polar reference frame in the HRN is inspired by a method for image registration, the Fourier-Mellin transform~\cite{casasent1976position,chen1994symmetric,reddy1996fft}. Image rotation and scaling correspond to translations in the log-polar reference frame. Thus, in the log-polar partition of the HRN, vector binding becomes equivariant to rotation and scaling. Going iteratively back and forth between the two reference frames, the HRN can successively infer the factors that correspond to these geometric transforms for a given input image.

Here, we extend the hierarchical resonator network to perform visual odometry. The generative model design of~\cite{renner2022T} is simplified to use a single image stored in working memory that acts as an allocentric map. The hierarchical resonator serves as a recursive filter that estimates four degrees of freedom of camera movement: two dimensions of translation, rotation, and scaling, by performing image registration between the current input and the map.
The first input image defines the (stationary) navigation coordinate frame, and the resonator network aligns subsequent images to it. After applying the transformation, the allocentric map can be updated dynamically by integrating the new sensor data. The map can also be designed to slowly forget content that is not refreshed. 
The full architecture is described in Fig.~\ref{fig:overview}. 
The full dynamics of the resonator are described in Eq.~\ref{eq:resonator_dynamics}, and the map update is shown in Eq.~\ref{eq:map_update2} in the Methods section.

We use input from an event-based camera (a Dynamic Vision Sensor, DVS~\cite{gallego2020event}). In our experiments, we generate images from the event stream -- so-called ``event frames'' -- by collecting a certain number of events. This is necessary as several events are required to perform reliable image registration. It takes around 7ms to collect the 2000 events for the shapes datasets. So, on average, the ``frame rate'' is more than 100 event-frames per second and increases adaptively when faster movement occurs as more events are generated.

In the VO setting, the camera is moving. Thus, the scene is dynamic, and the network receives a different input at each iteration. Consequently, the resonator network does not stay static but follows the input. 
Each successive input frame is typically quite similar to the previous frame. The resonator network utilizes its previous state information to solve the image registration for new images rapidly.

\subsection*{Experiments with the Event Camera Dataset} \label{sec:VO}
Following recent event-camera visual odometry work \cite{reinbacher_2017,rebecq2017real,zihao2017event,vidal2018ultimate,nguyen_2019,rebecq2019high,xiao2021research}, we use data from the \emph{Event Camera Dataset} \cite{mueggler2017event} to benchmark our architecture. The dataset contains events, frames, and IMU measurements from a DAVIS240C sensor and high-speed 6-DoF motion capture for the ground truth. First, we use the shapes\_rotation sequence, which is 60 seconds long and was recorded while the camera was rotated by hand in 3 DoF in front of a wall with images of geometrical shapes. The data sequence contains almost no translation but very fast rotation (angular velocity up to 730$^\circ$/s), which leads to motion blur and large jumps in the conventional camera frames but is well resolved by the DVS camera events. 

For simplicity, we assume that the camera movement happens only in the three rotational DoFs (roll, pan, and tilt) and approximate pan and tilt as pure translations in the pixel space. We show that the network tracks the camera movement faithfully with a median error of 3.5 degrees. As shown in Tab.~\ref{tab:literature_rotation}, the VSA-based resonator network outperforms the neural networks trained on samples of the shapes\_rotation dataset as reported by~\cite{nguyen_2019}. 
In addition, we tested the shapes\_translation sequence, where the camera was translated instead of rotated. Here, we assume the camera movement happens in four DoFs (x, y, z, and roll). The translation (x and y) is factorized in the Cartesian module of the HRN, and the rotation and scale (r and z) are factorized in the log-polar module of the HRN. For this dataset, as shown in Tab.~\ref{tab:literature_rotation}, neuromorphic VO network outperforms most neural networks reported by~\cite{nguyen_2019}.
Furthermore, we compare the error to other geometrical approaches that do not use neural networks in Tab.~\ref{tab:literature_translation}. 
For these algorithms, the relative position error is typically reported as a percentage of the total traveled distance. With an error of 0.53\%, the network, using only event data, performs similarly to the best visual-inertial odometry approaches. Note, \cite{vidal2018ultimate} reports better results (0.26\%) when including information from a frame-based sensor, but we only considered event-based methods for comparison.

\begin{table}[t]
\begin{tabular}{@{}lllcccl@{}}
\toprule
Publication  & Name 
 & \hspace{0.01cm}  & \begin{tabular}[c]{@{}l@{}} shapes rotation data \\ angle error ($^{\circ}$) \end{tabular} & \hspace{0.01cm}  &
\begin{tabular}[c]{@{}l@{}} shapes translation data \\ translation error ($m$) \end{tabular} & \\
\midrule
\cite{kendall_2015} Kendall et al. (2015) by \cite{nguyen_2019} & PoseNet  &  & 12.5 &  &  0.198 &  \\
\cite{kendall_2016} Kendall et al. (2016)  by \cite{nguyen_2019}& Bayesian PoseNet &  & 12.1 &  &  0.213  &  \\
\cite{laskar_2017} Laskar et al. (2017)  by \cite{nguyen_2019} &  Pairwise-CNN   &  & 10.4 &  & 0.225  &   \\
\cite{walch_2017} Walch et al. (2017)  by \cite{nguyen_2019}&  LSTM-Pose  &  & 7.6 &  & 0.108 & \\
\cite{nguyen_2019} Nguyen et al. (2019) &  SP-LSTM  & & 5 &  & 0.072  &   \\
This work & Neuromorphic VO &  & 3.5 (4.7)  &  & 0.078 &  \\
This work & Neuromorphic VO + IMU &  & 2.7 (4.2) &  & - &  \\
\bottomrule
\end{tabular}
\caption{\textit{Comparison of different neural network architectures based on the median angular error for the shapes\_rotation and the shapes\_translation recordings of the shapes dataset.} We report the median angle or translation error depending on the dataset, as calculated by \cite{nguyen_2019}. For the shapes\_rotation recording, the given numbers from our work are obtained using only the last part of the training set from~\cite{nguyen_2019} for calibration. In parentheses are the errors when the whole training set is used. For the shapes\_translation recording, the whole training set is used as there is insufficient movement in the z-axis.
}
\label{tab:literature_rotation}
\end{table}

Our approach does not require training; instead, we calibrate the trajectories by using part of the data after the experiment to find the lag and linear transform that minimize the error to the ground truth. For this calibration, we use either the same split (70/30) as~\cite{nguyen_2019} (numbers given in parentheses) or just the last 10 seconds of the same training set, which yields better results with less data. This is likely because it is closer to the test set and provides a larger movement range per time, excluding the slow movements at the beginning of the recording.

\begin{figure}[t]
  \centering
  \includegraphics[width=0.8\linewidth]{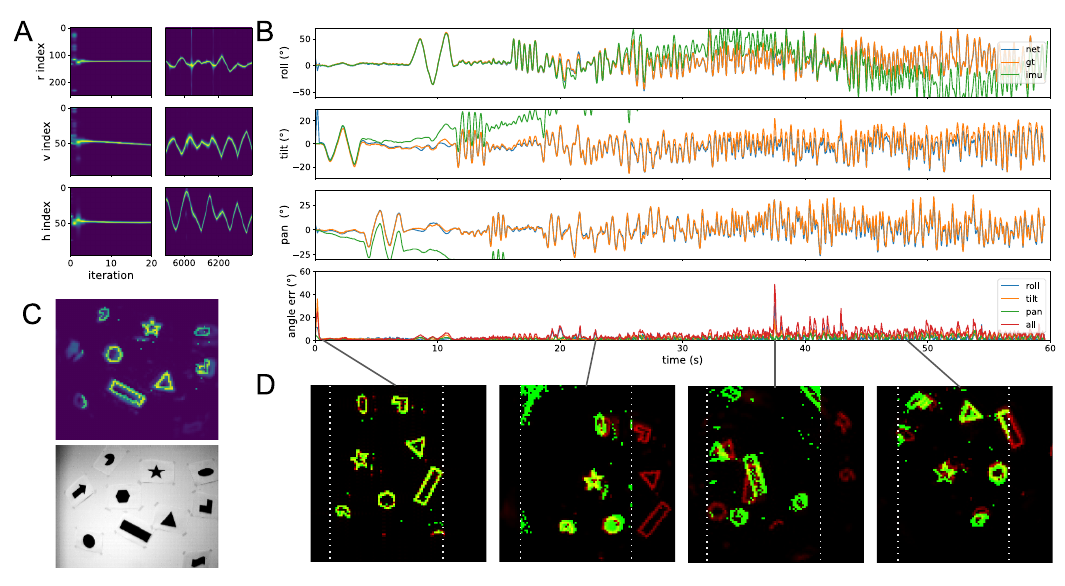}
  \caption{Tracking of the camera rotation from the event-based shapes\_rotation dataset \cite{mueggler2017event} in simulation. 
  \textbf{A.} Unprocessed readout (inner product similarity) of the resonator states at the beginning of the experiment (left) and in later iterations around second 37.5 of the dataset (right). The r index directly corresponds to the roll angle while v and h are rotated and scaled (calibrated) into pan and tilt angles. Brighter colors indicate a higher inner product similarity of the resonator state with the codebook vector at the given location.
  After a short orientation phase where several locations are active in superposition, the states converge to a unique solution and follow the camera's movement after less than ten iterations. In later iterations (right), we observe that the similarity peak also becomes broader (i.e., the network is less certain) when the camera moves quickly outside of the area covered by the map, leading to increased error (as seen in B and D). 
  \textbf{B.} Population vector readout (blue) of the angles from the HRN (Net) calibrated to the ground truth (gt) coordinates. Ground truth from motion capture (orange). Camera trajectory from IMU measurement (green). Lowest row: plot of the tracking error (angle between ground truth and output trajectory) over time.
  \textbf{C.} Map at the end of the experiment (top) and conventional camera frame that shows the scene in its entirety (bottom). 
  \textbf{D.} Event-based camera input (green) and transformed map (red) at 4 different iterations. Overlapping pixels between the map and camera view are yellow. The map was rotated and shifted by the current estimate of the network in order to show correspondence. White dotted lines indicate the borders of the camera image. The area outside of the white lines is zero-padded, as the transformed map can protrude the camera image.}
  \label{fig:results_shapes_rotation}
\end{figure}

The estimated camera trajectories compared to the ground truth are shown in Fig.~\ref{fig:results_shapes_rotation}. Fig.~\ref{fig:results_shapes_rotation}A shows the unprocessed readout of the resonator states over several iterations at the beginning and the middle of the experiment. The inner product similarity of the resonator state with the VSA encoding vectors that code for the given locations is plotted. The states are initialized randomly so that the similarity is approximately equally distributed over all locations (left panels) in the beginning. In an orientation phase, in the first few iterations, the resonator quickly identifies several possible solutions that are active in superposition before it starts tracking the correct camera trajectory a few iterations later. 

The network output, after calibration, and the ground truth match closely~(Fig.~\ref{fig:results_shapes_rotation}B). The trajectory obtained from integrating (dead reckoning) the readings from the inertial measurement unit (IMU), which is also shown in Fig.~\ref{fig:results_shapes_rotation}B, drifts strongly, as expected from dead reckoning. In our VO network, drift is considerably smaller, as the movement is tracked relative to the map without dead reckoning. 

The dynamic map at the end of the experiment is shown in Fig.~\ref{fig:results_shapes_rotation}C. Fig.~\ref{fig:results_shapes_rotation}D shows the input to the network (green) and the map transformed to the egocentric coordinates for visualization purposes at several points during the experiment. The first image is taken directly after the orientation phase when the overlap between the map and the input is perfect. The third image is taken around the iteration with the largest error. A likely reason for the increased error in this iteration is a combination of poor correspondence between the map and the current input and very fast camera movements that make it hard to follow with the slower state update dynamics.

\begin{table}[ht]
\begin{tabular}{@{}llcllllll@{}}
\toprule
Publication  & Name & rel. position error ($\%$) & &  \\
\midrule
\cite{zihao2017event} Zhu et. al. 2017 & EVIO (E+I) & 2.42 &  &   \\
\cite{rebecq2017real} Rebecq et al. 2017  &  Real-time visual-inertial odometry (E+I)  & 0.5 &  &   \\ 

\cite{vidal2018ultimate} Vidal et al. 2018  & Ultimate SLAM (E+I)  & 0.41 &  & \\

\cite{xiao2021research} Xiao et al. 2021 &  (E+I) & 0.45 &  &  \\
This work & Neuromorphic VO (E) & 0.53 &  &   \\
\bottomrule
\end{tabular}
\caption{
\textit{Comparison of the relative position error as a percentage of distance traveled for the shapes\_translation dataset.} Different publications use different modalities in their VO approaches. They use events (E) and inertial sensor data (I). We have not included methods that use regular camera frames. 
}
\label{tab:literature_translation}
\end{table}

\subsection*{Sensory fusion of IMU and event-based vision} \label{sec:fusion}
\begin{figure}[t]
  \centering
  \includegraphics[width=0.5\linewidth]{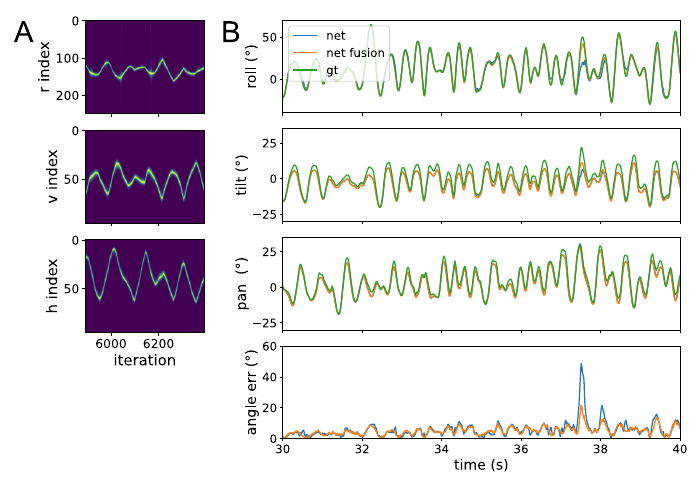}
  \caption{Tracking of the camera rotation from the event-based shapes\_rotation dataset \cite{mueggler2017event} in simulation using both IMU and event-based vision sensors.
  \textbf{A.} Unprocessed readout of the resonator states around second 37.5 of the dataset, for the network including IMU fusion, compare with Fig.~\ref{fig:results_shapes_rotation}A.
  \textbf{B.} Comparison of the trajectories and error with and without fusion. The network that uses both VO and IMU performs better in cases where VO is difficult.
  }
  \label{fig:results_fusion}
\end{figure}

To improve the network's performance when VO generates a large error and to emphasize the flexibility of our approach and the VSA framework, we demonstrate sensory fusion of the visual and inertial modalities in Fig.~\ref{fig:results_fusion}.  While the visual odometry provides absolute angles relative to the map, the IMU measurements provide the angular velocities, i.e., the rates of change of the rotation angles.
This means we can include a prediction step into the state update equations (see Eq.~\ref{eq:fusion}). Instead of just estimating each state from the input and the other states, each state is predicted using its approximate rate of change before it is used to update the other states. 
As expected, the peak in the readout becomes sharper in Fig.~\ref{fig:results_fusion}A compared to Fig.~\ref{fig:results_shapes_rotation}A, and particularly in difficult iterations, the error is notably reduced, leading to an overall lower median error in this experiment of 2.7 degrees (see Tab.~\ref{tab:literature_rotation}). 

\subsection*{Robust VO with a dynamic scene} \label{sec:robot}
\begin{figure}[t]
  \centering
  \includegraphics[width=0.9\linewidth]{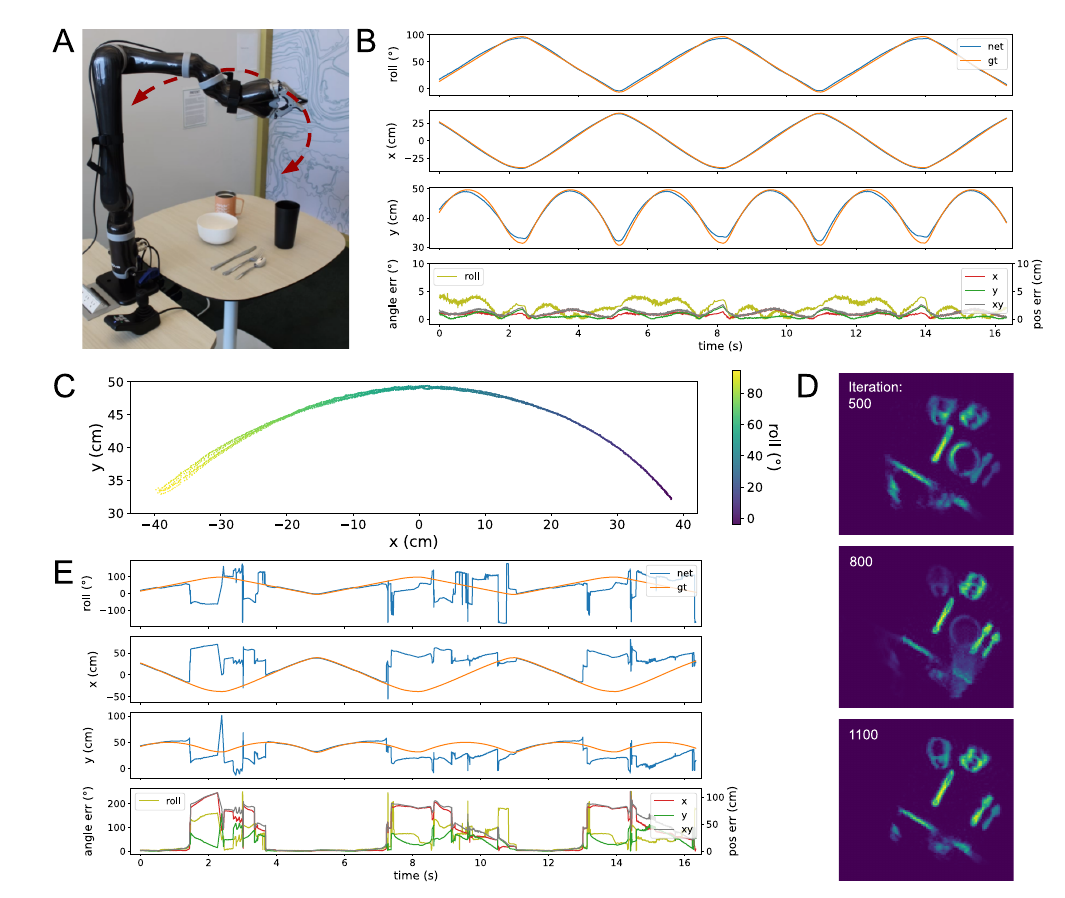}
  \caption{Tracking of the location and rotation of an event-based camera mounted on a robotic arm. 
  \textbf{A.} The robotic arm setup and the tabletop scene with the event-based camera mounted on the arm. The arm moves back and forth in the arc shown in red. 
  \textbf{B.}  Population vector readout (blue) of the angles transformed to the ground truth coordinates. Ground truth from the robotic arm (orange). Lowest row: tracking error over time for location (left axis) and rotation (right axis).
  \textbf{C.} Population vector readout of the roll angle (color) and x, y locations transformed into the ground truth coordinate system.
  \textbf{D.} Tracking and mapping are robust against small changes in the scene. The map at the iteration before removal of the bowl, just after removal, and seconds after removal. As soon as the bowl is removed, it fades from the map until it is fully deleted.
  \textbf{E.} Same as B. but without map learning. When the arm moves out of the range of the initial map (near 2s, 8s, and 14s), the tracking no longer works. However, the network can recover when the map comes back into view. The trajectory is aligned the same way as B.
  }
  \label{fig:results_robot_arm}
\end{figure}

To demonstrate the independence of the algorithm on the specific dataset and camera and to allow us to record a controlled dynamic scene, we built a custom setup using a robotic arm with an event-based camera mounted on the wrist. The arm was moved repeatedly in an arc over a static tabletop scene, as shown in Fig.~\ref{fig:results_robot_arm}A. The network tracks the camera's movement relative to the starting position in three degrees of freedom (horizontal and vertical translation and roll angle). Here, the network achieves a median error of 1.7 cm in the position and 1.5 degrees in the roll angle. The estimated trajectories, the comparison with the ground truth, and the errors are shown in Fig.~\ref{fig:results_robot_arm}B and C. 

For simplicity, our system assumes that scenes are planar and mostly static, i.e., the objects do not move. However, in reality, this assumption rarely holds. In future work, it is possible to include separate moving 3D objects in the generative model of the scene. However, in this proof of concept, we content ourselves with testing the robustness of the VO algorithm to dynamic changes in the scene by manually removing a large object, e.g., the bowl in Fig.~\ref{fig:results_robot_arm}A. Removal of the bowl does not increase the error notably (by $0.2 ^\circ$ in angle and no increase in position), and we visualize in Fig.~\ref{fig:results_robot_arm}D how the bowl vanishes from the dynamic map throughout the experiment. 

The dynamic map update lends the network robustness to dynamic scene changes, and it allows the network to operate in regions of the scene not captured in the initial map. In an ablation experiment of the map update (Fig.~\ref{fig:results_robot_arm}E), we demonstrate that the network fails to track the camera motion when the map update is disabled and the initial map's content moves out of view. We also observe that the network recovers once the familiar input is in view again.

While the map update is advantageous, in longer simulations, there is a risk that the map (and, therefore, our navigation coordinate frame) drifts because the estimated transforms used for the map updates are not always accurate. To avoid such drift, we anchor the map by adding the initial map at each map update, as described in Eq.~\ref{eq:map_update2}. This makes the initial map a fixed point of the dynamics, i.e., without input, the map decays to this initial map. The effect of this anchoring can be seen in Fig.~\ref{fig:results_robot_arm}C. The position variance is increased when the arm moves to the left, which is further away from the initial map. In future work, mechanisms for re-anchoring with a hierarchy of maps could be explored.

\section*{Discussion}  
We propose a novel solution to visual odometry (VO), or self-motion estimation -- one of the fundamental tasks in robotics \cite{nister2004visual,scaramuzza2011visual}. On a data set collected with an event-based camera \cite{mueggler2017event}, we achieve state-of-the-art performance of the proposed VO algorithm. In a simplified setting, we also demonstrate its applicability to a robotics task. While more research is required to expand this approach to 3D movement and scenes, this is the first demonstration of the feasibility of using VSA resonator networks for VO.

In the last decade remarkable progress has been made in performing visual odometry tasks with event-based cameras \cite{cook_2011,kim_2014,mueggler_2014,censi_2014, weikersdorfer_2014, gallego_2015,kueng_2016, rebecq_2017EMVS, rebecq2016evo, gallego2018unifying, zhu_2019}. 
The use of event-based cameras enables high temporal resolution ($>1kHz$), robustness to light intensity changes, with reduced motion blur at a lower computational and energy cost compared to high-speed cameras.
Efficient solutions to VO are crucial for energy-constrained robotics applications such as small autonomous drones~\cite{palossi_2019} and planetary rovers~\cite{moravec1980obstacle, lacroix1999rover, corke2004omnidirectional}.

Our approach leverages VSAs, recently proposed as a framework for programming neuromorphic hardware~\cite{KleykoComputingParadigm2021}. Notably, the proposed solution combines three recent innovations in VSAs: (1) fractional power encoding \cite{frady_2019precession, komer_2019, frady2021VFA} for encoding continuous quantities and transforms, (2) resonator networks \cite{frady_2020_resonator,kent_2020resonator} for efficiently analyzing the visual input, and (3) phasor-based VSA models \cite{frady_2019tpam, frady_2019precession} for enabling a spiking neuromorphic implementation. 

Neuromorphic computing holds great promise for efficiency, as demonstrated in several applications~\cite{davies2021advancing}. Designing algorithms for this unconventional computing architecture, however, is a challenge. 
Our results showcase a new VSA-based computing framework for programming neuromorphic computations. 
As shown in the companion paper~\cite{renner2022T}, the algorithm can be directly converted into a spiking implementation, which allows us to implement the resonator network onto Intel's research chip Loihi \citep{davies2018loihi}.
Further details of this implementation and demonstration of gains in power efficiency are presented there~\cite{renner2022T}.

Currently, most researchers use deep-learning artificial neural networks as a programming framework for neuromorphic chips. This framework parameterizes algorithms with artificial neurons configured in layered, densely connected, feed-forward networks~\cite{EsserEtAl2016, frenkel20180, shrestha2018slayer, renner2021backpropagation}.
Unsurprisingly, feed-forward processing with artificial neurons underuses neuromorphic architectures to yield only modest gains over implementations in conventional computers or GPUs~\cite{davies2021advancing}. It is the exploitation of spike-timing, intrinsic neural dynamics, and recurrent architectures~\cite{davies2021advancing, davies2018loihi, frady_2020knn} that yield orders of magnitude benefits to energy-delay products in neuromorphic hardware~\cite{davies2021advancing}. While we present results in CPU simulations here, our architecture expresses all of the above features when converted to a spiking neuromorphic implementation. These aspects make the design promising for fully realizing the benefits of neuromorphic devices.

One challenge we have observed with deep learning approaches to VO is overfitting the specific scene(s) used in training. 
In the benchmark experiments, we used for comparison, training and test sets were extracted from the same event sequence~\cite{nguyen_2019}.
In contrast, our model holds the visual scene in the working memory of the dynamic map, allowing for generalization over all possible scenes, unlike in convolutional neural network approaches~\cite{serranogotarredona_2009}. 
We also show that we can easily update the allocentric map for tracking, as demonstrated in the map update experiment.
The network also has the ability to recover in cases when tracking is lost temporarily.

Furthermore, the resonator network is small compared to the convolutional approaches while performing similarly or better. The convolutional networks that these methods use as a central part of their architectures have 4-138 million trainable parameters and use several billion operations per sample (\cite{nguyen_2019}: VGG16, \cite{kendall_2015,kendall_2016,walch_2017}: GoogLeNet, \cite{laskar_2017}: ResNet34). The HRN, on the contrary, only has a few dozen parameters to adjust the behavior of the network and for calibration and requires around 360 million operations per iteration, as a crude estimate, 1-2 orders of magnitude less than the compared networks.

Computationally, our approach is related to previous work on spatial correlation-based approaches, particularly the map-seeking circuit (MSC)~\cite{arathorn2002map}. The map-seeking circuit is a multi-layer network with feedback that performs inference in generative models and has been successfully used for image registration in adaptive optics~\cite{vogel2006retinal}. While these approaches share superposition and content-addressable memory for estimating the transforms, we avoid representing the transforms with large sets of matrices by making use of VSA vector binding.

In future work, the model could be augmented using multi-headed resonator networks \citep{renner2022T} for more detailed scene analysis and VO. This network design could be used to recognize and track multiple individual objects, which could be a step toward ``semantic SLAM''~\cite{bowman2017probabilistic}.
This could also make the system robust to objects that move. 

While the model validates the VSA and hierarchical resonator approach in a robotics task, it is still a limited demonstration of the framework's capability, as we considered a 2-dimensional scene in our robotic setup and the dataset benchmark and because we perform VO only with 3 (three rotations) or 4 (translation and roll) degrees of freedom. 
Scaling this architecture to a fully-fledged 3D scenario with a 6-degree-of-freedom motion estimate requires further extensions of the hierarchical resonator theory and likely the inclusion of learning and additional modalities. 
We have also shown that integrating additional sensor channels, such as an inertial measurement for motion estimation, leads to more effective and precise multi-modal odometry. In future work, this approach could be extended with depth perception, e.g., stereo or LIDAR technology. The resonator framework and VSA representations are well-suited for fusion of different sensory modalities due to their use of homogeneous vector-based representations for different spaces and transformations between them.

\section*{Methods}

\subsection*{Data and preprocessing}

\paragraph{Event camera dataset.}
The event-based dataset \cite{mueggler2017event} that we used in our validation experiment contains events and IMU  measurements from a DAVIS240C camera~\cite{brandli_2014} and high-speed 6-DoF motion-capture for the ground truth. The shapes\_rotation sequence is 60 seconds long and was recorded holding the camera in hand and rotating it in 3 rotational degrees of freedom. It contains almost no translation but very fast rotation (angular velocity up to $730~^\circ/s$). The shapes\_translation sequence is 60 seconds long and was recorded holding the camera in hand and translating it in 3 degrees of freedom. It contains only a small amount of rotation but fast translation (velocity up to $2.6~m/s$).\\

\paragraph{Robotic arm setup.} \label{sec:arm}
We ran another set of experiments to demonstrate VO with an event-based camera mounted on a robotic arm.
We used a Prophesee event-based camera evaluation kit for the experiment: an event-based vision sensor with VGA resolution (640x480),  a C mount lens (70 degrees FOV), and Prophesee player software. A Prophesee Gen2 Metavision event-based camera was mounted on a Jaco2 robot arm - specifically, on the joint that meets the end-effector - using a wristband (see Fig.~\ref{fig:results_robot_arm}). The Jaco2 arm is a 6DoF assistive robot arm.
The arm is mounted on a table with several objects, including a cup, utensils, and bowl, placed on top to simulate real-world use cases (see Fig.~\ref{fig:results_robot_arm}A). The robot arm's reach is 90 cm, and it is mounted on the platform next to the table of size 75cm x 75cm. 

We moved the arm in the 2D plane with the camera facing down, scanning the table. We recorded the data using the Prophesee player and converted the events to NumPy arrays using Prophesee's metavision package 2.1.0. We used the Kinova Jaco2 and custom-built software to control the arm and record its trajectory. \\

\paragraph{DVS data preprocessing}
For all network simulations, recorded events were used. Events were stored with a timestamp,  pixel coordinates (x, y), and a polarity (on/off).
Polarity was discarded. Coordinates were downsampled from 640x480 to 64x48 pixels for the Prophesee sensor and from 240x180 to 96x72 pixels for the INIvation DAVIS240C. This was achieved by dividing the event coordinates by the respective factor and then rounding them to integers. For the shapes\_rotation dataset, a fixed amount of 2000 events, and for the robotic arm dataset, 5000 events were accumulated into one event package (set of events), which makes the processing mostly invariant to the current movement speed. Especially for faster movements, even smaller packages would be possible and yield a better temporal resolution. Finally, a binary array $I$ was created by setting all pixels with more than 0 (shapes\_rotation) or 1 (robotic arm) events to 1 and all others to 0, effectively removing some noise events in the latter case.

\subsection*{Encoding}
Throughout this paper, we use the Fourier Holographic Reduced Representations (FHRR) VSA~\cite{ plate_1994,plate1995holographic}, which uses vectors of complex numbers like the Fourier transform.  We use fractional power encoding (FPE)~\cite{plate_1994, frady_2019precession, komer_2019, frady2021VFA} to represent space, as described in \cite{renner2022T}. In this vector encoding, space is spanned by repeated binding of a VSA “seed vector,” while the binding operation is generalized to fractions so that it can be applied on a continuous scale~\cite{plate_1994, frady_2019precession, komer_2019, frady2021VFA}.
Instead of synthetic images of letters, as in \cite{renner2022T}, we encoded an event package into a VSA vector.
With FPE, a set of events $E$ can be encoded into a vector $s$ as the sum of the exponentiated seed vectors $h_0$ and $v_0$ (for horizontal and vertical coordinates, respectively):

\begin{align}
    \mathbf{s} &= \sum_{x,y \text{ in } E}{ \mathbf{h}_0^{x} \odot \mathbf{v}_0^{y} }, \label{eq:evtenc_sum}
\end{align}

Here, the sum denotes bundling, which is the sum of complex numbers, and $\odot$ signifies binding, which is the element-wise (Hadamard) product.
For instance, a set of 3 events  ($E = \{ [27,24], [39,43], [28,38]\}$) is encoded as follows:

\begin{align}
    \mathbf{s} &= \mathbf{h}_0^{27} \odot \mathbf{v}_0^{24}
    + \mathbf{h}_0^{39} \odot \mathbf{v}_0^{43} 
    + \mathbf{h}_0^{28} \odot \mathbf{v}_0^{38}. \label{eq:evtenc_example}
\end{align}

In practice, as long as the possible ranges of x and y are known beforehand, which is the case for camera pixels, the encoding can also be written as a matrix product of the complex codebook matrix $\Phi$ and the binary event array I. The codebook matrix contains a codebook vector for each pixel location $[x,y]$ that is calculated by $ \mathbf{h}_0^{x} \odot \mathbf{h}_0^{y}$. In this formulation, the event package can be encoded by:

\begin{align}
    \mathbf{s} &= \Phi I \label{eq:evtenc_prod}.
\end{align}

The FPE encoding constitutes a generalization of the Fourier transform. In the case of regularly spaced seed vector phases containing all N $N^{th}$-roots of unity, the 1D FPE becomes equivalent to a 1D Fourier transform. In the 2D case described above, $\Phi$ would be the discrete Fourier transform (DFT) matrix.
This can be used to achieve equivariance of the binding operation with periodic transformations, such as rotation.
A similar mechanism to encode event-based vision pixels into a hyperdimensional vector was employed by \cite{mitrokhin_2019} using permutations instead of binding with the Hadamard product.
The error reported for the shapes\_rotation dataset, and the trajectories shown in Fig.~\ref{fig:results_shapes_rotation} were obtained using the regular DFT codebook matrix. The trajectory in Fig.~\ref{fig:results_robot_arm} was obtained using a random codebook matrix with a vector size of N = 3072.

\subsection*{Network architecture}
As shown in Fig.~\ref{fig:overview}, the network can be split into three parts: (1) The hierarchical resonator that performs the tracking, (2) the reference frame transformation from the camera to the map coordinates, and (3) the map update.
The hierarchical resonator receives the encoded image and the map as input and outputs estimates of the transformation between the two inputs.
The next module uses these estimates to transform the encoded input image from the camera to map coordinates. This transformed image in map coordinates is then used to update the map.

The resonator network was introduced in \cite{frady_2020_resonator}. 
Here, we use the hierarchical resonator network \cite{renner2022T} that generalizes the resonator by allowing a nested structure of factors that can use different reference frames (such as log-polar coordinates).
The version used here is simpler than in \cite{renner2022T}, as we do not include color and scale. Furthermore, we only have a single template, which we call a map, in this paper. However, unlike in \cite{renner2022T}, the map changes over time, and the input to the network is not a static artificial letter but a stream of events from a DVS camera.

\begin{figure}[!ht]
  \centering
  \includegraphics[width=0.5\linewidth]{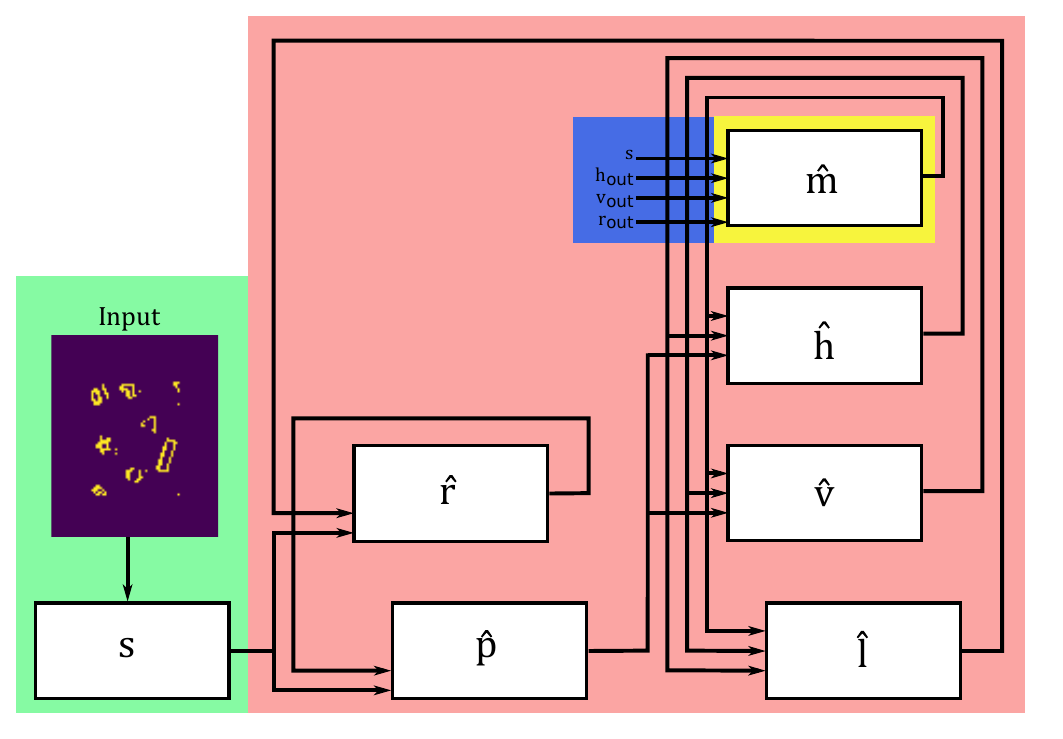}
  \caption{\textit{Hierarchical resonator for visual odometry.} Colors match Fig.\ref{fig:overview}. The dynamics are explained in Eq.\ref{eq:resonator_dynamics}.}
  \label{fig:resonator_vo}
\end{figure}

The evolution of the three states of the resonator network (corresponding to horizontal displacement $h$, vertical displacement $v$, and rotation $r$) is described by the following equations and visualized in~\ref{fig:resonator_vo}. This describes the HRN used for the shapes\_rotation dataset and the robotic arm data. For the shapes\_translation dataset, we add a fourth factor for scale that accounts for camera movement in the z-axis.

\begin{align}
\begin{split}
    \mathbf{\hat{h}}(t+1) &= (1-\gamma) \mathbf{\hat{h}}(t) + \gamma f \Big( \mathbf{H}p\Big(\mathbf{H}^\dagger \Big(\mathbf{\hat{p}}(t) \odot \mathbf{\hat{v}}^*(t) \odot \mathbf{\hat{m}}^*(t)  \Big) \Big) \Big), \\
    \mathbf{\hat{v}}(t+1) &= (1-\gamma) \mathbf{\hat{v}}(t) + \gamma f \Big( \mathbf{V}p\Big(\mathbf{V}^\dagger \Big(\mathbf{\hat{p}}(t) \odot \mathbf{\hat{h}}^*(t) \odot \mathbf{\hat{m}}^*(t)  \Big) \Big) \Big), \\
    \mathbf{\hat{r}}(t+1) &=  (1-\gamma) \mathbf{\hat{r}}(t) + \gamma f \Big( \mathbf{R}p\Big(\mathbf{R}^\dagger \Big( \mathbf{s}(t) \odot \mathbf{\hat{l}}^*(t) \Big) \Big) \Big), \\
    \text{with}\\
    \mathbf{\hat{p}}(t) &=  \mathbf{\Lambda} \big( \mathbf{s}(t) \odot \mathbf{\hat{r}}^*(t) \big), \\
    \mathbf{\hat{l}}(t) &=  \mathbf{\Lambda}^{-1} \big( \mathbf{\hat{m}} \odot \mathbf{\hat{h}}^*(t) \odot \mathbf{\hat{v}}^*(t)  \big).
\end{split}
\label{eq:resonator_dynamics}
\end{align}
$\mathbf{s}(t)$ is the input as obtained from Eq.~\ref{eq:evtenc_prod}. The letters with the $\hat{}$ are the current states of the resonator at a given iteration t. The capital letters ($H, V, R$) are the encoding matrices for the respective states. The encoding matrices are created by generating a codebook vector for each location using FPE. Cleanup of the state is achieved by first decoding the state with the complex conjugate transpose ($\dagger$) of the encoding matrix, taking the real value and applying an optional nonlinearity, and then encoding it again by a matrix product with the encoding matrix.
$f$ denotes a nonlinearity, specifically, elementwise division of the complex vector by the complex magnitudes, i.e., projection to the unit circle in the Gaussian number plane (phasor projection).  $p$ is another nonlinearity that leads to a stronger cleanup (e.g., ReLU, exponentiation, and normalization; for details, refer to \cite{renner2022T}). $\Lambda$ denotes the matrix that transforms between the VSA vector in Cartesian coordinates and the one in polar or log-polar coordinates. $\gamma$ is constant and determines the speed of the state update. $*$ denotes the complex conjugate necessary for unbinding of a state. Note that if binding corresponds to a shift in one direction, unbinding corresponds to the same shift but in the opposite direction.
The codebooks used for the readout and cleanup and that define the resonator's attractor dynamics are not necessarily the same as the ones used for encoding the image; they just need to be created from the same FPE seed vector that represents the 'unit' shift in a particular dimension. In this work, we make use of this in two cases: First, for the shapes\_translation resonator with 4 DoF, we constrain the possible solution space for scales and roll rotations by only including the expected scales and angles. Second, we improve the spatial resolution of the scale factor by including fractional scales (fractional powers in the FPE~\cite{frady2021VFA}) that do not correspond to pixels but rather to the space in between pixels in the log-polar image. This effectively creates a continuous attractor in the factor's dynamics, which will be studied further in future work.

The readout of the resonator is done as follows: First, the state is decoded using the decoding matrix. This corresponds to calculating the inner product similarity of the resonator state with the codebook vectors at each location of interest. To get the best locations, one could take the index with the largest similarity value. This, however, discards much of the information contained in the similarity distribution (see Fig.~\ref{fig:results_shapes_rotation}A and Fig.~\ref{fig:results_fusion} for plots of this unprocessed state readout). Therefore, the output of the network is calculated as a population vector of the state readout (similar to \cite{renner_2019}). The population vector is the similarity-weighted average of the indices (see Eq.~\ref{eq:popvec}). It allows sub-pixel (sub-index) resolution compared to simply taking the index of the largest value. To avoid the influence of outliers, only the five neighboring indices on both sides of the index with the largest value were used for the population vector decoding. I.e., first, the highest peak in the readout is selected, and then the neighborhood around the peak is used to determine the exact estimate.

For instance, for the horizontal displacement (h) state, the network output ($h_{out}$, and analogously $v_{out}$ and $r_{out}$) is calculated as follows:
\begin{align}
& h_{sim}(t) = H^\dagger\mathbf{\hat{h}}(t), \\
& h^{max}_{sim} = argmax(h_{sim}(t)), \\
& h_{out}(t) = \frac{\sum_{i=[h^{max}_{sim}-5, h^{max}_{sim}+5]}{i  h_{sim,i}(t)}}{\sum_{i=[h^{max}_{sim}-5, h^{max}_{sim}+5]}{h_{sim,i}(t)}}. 
 \label{eq:popvec}
\end{align}

The map estimate $\hat{m}$ is treated almost like a regular resonator state. However, a cleanup matrix does not exist, as the map is unknown beforehand and changes over time. Unlike the other states, which are initialized by a random complex vector of unit magnitude, the first map vector $\hat{m}(0)$ is the first input $s(0)$ to the network. This means that all transformations will be relative to this starting position, and the origin of the map reference frame is set to be at the origin of this first map. After that, the map is updated (Eq.~\ref{eq:map_update2}) similarly to the other states by an estimate created from the other states. This estimate is the input rotated and shifted from input to map coordinates, see Eq.~\ref{eq:map_update1}). 
This transformation could, in principle, be performed using the cleaned-up resonator states. However, to achieve a cleaner map, the update is performed using the output of the network, i.e., only with the best transformation estimate (i.e., the network output as described in Eq.~\ref{eq:popvec}) instead of a superposition of all likely transformations as needed for the resonator.  

As described in Eq. \ref{eq:map_update1}, the transformation of the map is performed using fractional binding of the current input vector $s(t)$ with the seed vectors exponentiated by the real-valued output of the network.  

To avoid a wrong map update, which could lead to a remapping to a different frame of reference, the map update is blocked during the first 100 iterations (after which the correct transformation has usually been found). 

To avoid drifts of the map that can happen if the estimates are slightly off over a longer time, the map is anchored to the starting map. This is achieved by adding the starting map $\hat{m}(0)$ to the map update at each iteration with a low weight $\mu_2$. Here, the anchor is the first map that defines the navigation frame and does not change; however, the anchor could also be implemented as a slowly updating long-term memory.

\begin{align}    
    \mathbf{m}(t) &=  \big[ \mathbf{\Lambda} \big( \mathbf{s(t)} \odot \mathbf{h}^{h_{out}(t)} \odot \mathbf{v}^{v_{out}(t)}  \big) \big] \odot  \mathbf{r}^{r_{out}(t)},
    \label{eq:map_update1} \\
    \mathbf{\hat{m}}(t+1) &= \mu_1 \mathbf{\hat{m}}(t) + \mu_2 \mathbf{\hat{m}}(0) + (1-\mu_1-\mu_2) \mathbf{m}(t).
    \label{eq:map_update2}
\end{align}

For the IMU fusion experiments, the states are additionally updated before every resonator iteration using fractional binding with the IMU reading from the shapes\_rotation dataset, interpolated to the current timestamp, and adjusted to the time difference between the current and the last timestamp.

For instance, the roll angle estimate $\hat{r}$ is updated as follows:
\begin{align}  
\mathbf{\hat{r}}(t) = \mathbf{\hat{r}}(t-1) \odot r^{r_{IMU}(t)}, 
\label{eq:fusion}
\end{align}
where $r$ is the seed vector for the roll angle and $r_{IMU}(t)$ is the IMU reading interpolated at time t.

\subsection*{Analysis}

For the comparison with the ground truth, we obtained the raw state trajectory over time by decoding the states by the population vector readout explained above for each iteration. Each iteration was assigned the timestamp in the middle of the first and last event of the respective event package processed in that iteration. To determine the lag, we resampled the ground truth and the raw output trajectory to a fixed sampling rate of 400 Hz using linear interpolation. Finally, we calculated the lag between the ground truth and the network by cross-correlation of the camera roll trajectory.  

To determine the error, the ground truth trajectories were resampled using linear interpolation to the timestamps of the network trajectories.
A part of the trajectories was used to find the best matching scaling, translation, and rotation using the Umeyama algorithm for rigid alignment \cite{umeyama1991least} to calibrate the pan and tilt (or horizontal and vertical) trajectories. For the shapes\_rotation dataset, we used either the same 70/30 dataset split (called "novel split") as \cite{nguyen_2019} or we only used the last 10 seconds of the training set for calibration. We used the first arc ($0.5-5s$) for the robotic arm data. This calibration is needed as the ground truth reference frame is not necessarily the same as the first map reference frame. Also, the network calculates the translation in pixel coordinates.
The roll angle was set to start at 0 but was not calibrated as the network output is in absolute angles like the ground truth. 
Note that the calibration maps the network's v and h output traces to pan and tilt angles in case of the shapes\_rotation data and to x and y positions in case of the robotic arm data. 
The results reported for the shapes\_translation dataset in Table \ref{tab:literature_translation} are obtained using the trajectory evaluation method and code from \cite{Zhang18iros}. The trajectories were calibrated beforehand to align the scale factor to the movement in the z-axis. Then the similarity transformation `sim3' from the trajectory evaluation repository is used for alignment.

Preprocessing and simulations were conducted using NumPy (1.22.3) in Python (3.9), and plots were generated with matplotlib. We used SciPy (1.7.3) for interpolation and the Python package quaternion (2022.4.2) for processing the ground truth trajectory of the shapes dataset.

\section*{Data availability}
The event-based shapes dataset \cite{mueggler2017event} is publicly available at \href{https://rpg.ifi.uzh.ch/davis_data.html}{https://rpg.ifi.uzh.ch/davis\_data.html}. The robotics arm data generated and analyzed during the current study is available on CodeOcean~\cite{renner_2024_codeocean_vo}.

\section*{Code availability}
The source code to demonstrate the hierarchical resonator on the VO task~\cite{renner_2024_codeocean_vo} is available on CodeOcean at \href{https://doi.org/10.24433/CO.6568112.v1}{https://doi.org/10.24433/CO.6568112.v1}.

\section*{Acknowledgments}
A.R. thanks his former students Céline Nauer, Aleksandra Bojic, Rafael Peréz Belizón, Marcel Graetz, and Alice Collins for helpful discussions. Y.S. and A.R. disclose support for the research of this work from the Swiss National Science Foundation (SNSF) [ELMA PZOOP2 168183]. A.R. discloses support for the research of this work from Accenture Labs, the University of Zurich postdoc grant [FK-21-136] and the VolkswagenStiftung [CLAM 9C854]. 
F.T.S. discloses support for the research of this work from NIH [1R01EB026955-01] and NSF [IIS2211386].

\section*{Author Contributions Statement}

A.R., L.S., A.D., G.I., E.P.F., F.T.S., and Y.S. contributed to writing and editing of the manuscript; Y.S. and L.S. conceptualized the project in the robotic space; A.R., E.P.F. and F.T.S. conceptualized the project in the algorithmic space; A.R. developed the VO network model, performed, and analyzed the network simulations; L.S. performed the robotic arm experiments.


\begin{thebibliography}{10}
\providecommand{\url}[1]{{#1}}
\renewcommand{\doi}[1]{\href{https://doi.org/#1}{doi:#1}}

\bibitem{Srinivasan2010bees}
M.~Srinivasan, Honey bees as a model for vision, perception, and cognition.
\newblock Annual Review of Entomology \textbf{55}, 267--284 (2010)

\bibitem{nister2004visual}
D.~Nist{\'e}r, O.~Naroditsky, J.~Bergen, \emph{Visual odometry}, in
  \emph{Proceedings of the 2004 IEEE Computer Society Conference on Computer
  Vision and Pattern Recognition, 2004. CVPR 2004.}, vol.~1 (IEEE, 2004), pp.
  I--I

\bibitem{scaramuzza2011visual}
D.~Scaramuzza, F.~Fraundorfer, Visual odometry [tutorial].
\newblock IEEE robotics \& automation magazine \textbf{18}(4), 80--92 (2011)

\bibitem{palossi_2019}
D.~Palossi, A.~Loquercio, F.~Conti, E.~Flamand, D.~Scaramuzza, L.~Benini, A
  64-{mW} {DNN}-based visual navigation engine for autonomous nano-drones.
\newblock {IEEE} Internet of Things Journal \textbf{6}(5), 8357--8371 (2019).
\newblock \doi{10.1109/JIOT.2019.2917066}.

\bibitem{moravec1980obstacle}
H.P. Moravec, Obstacle avoidance and navigation in the real world by a seeing
  robot rover.
\newblock Ph.D. thesis, Stanford University (1980)

\bibitem{lacroix1999rover}
S.~Lacroix, A.~Mallet, R.~Chatila, L.~Gallo, \emph{Rover self localization in
  planetary-like environments}, in \emph{Artificial Intelligence, Robotics and
  Automation in Space}, vol. 440 (1999), p. 433

\bibitem{corke2004omnidirectional}
P.~Corke, D.~Strelow, S.~Singh, \emph{Omnidirectional visual odometry for a
  planetary rover}, in \emph{2004 IEEE/RSJ International Conference on
  Intelligent Robots and Systems (IROS)(IEEE Cat. No. 04CH37566)}, vol.~4
  (IEEE, 2004), pp. 4007--4012

\bibitem{mead_1990}
C.~Mead, Neuromorphic electronic systems.
\newblock Proceedings of the {IEEE} \textbf{78}(10), 1629--1636 (1990).
\newblock \doi{10.1109/5.58356}.

\bibitem{chicca_2014}
E.~Chicca, F.~Stefanini, C.~Bartolozzi, G.~Indiveri, Neuromorphic electronic
  circuits for building autonomous cognitive systems.
\newblock Proceedings of the {IEEE} \textbf{102}(9), 1367--1388 (2014).
\newblock \doi{10.1109/JPROC.2014.2313954}.

\bibitem{davies2021advancing}
M.~{Davies}, A.~{Wild}, G.~{Orchard}, Y.~{Sandamirskaya}, G.A.F. {Guerra},
  P.~{Joshi}, P.~{Plank}, S.R. {Risbud}, Advancing neuromorphic computing with
  loihi: A survey of results and outlook.
\newblock Proceedings of the IEEE pp. 1--24 (2021).
\newblock \doi{10.1109/JPROC.2021.3067593}

\bibitem{Indiveri_Liu_2015}
G.~Indiveri, S.-C.~Liu, Memory and information processing in neuromorphic
  systems.
\newblock Proceedings of the IEEE \textbf{103}(8), 1379--1397 (2015).
\newblock \url{https://arxiv.org/pdf/1506.03264.pdf}

\bibitem{PlateAnalogy2000}
T.A. Plate, {Analogy Retrieval and Processing with Distributed Vector
  Representations}.
\newblock Expert Systems: The International Journal of Knowledge Engineering
  and Neural Networks \textbf{17}(1), 29--40 (2000)

\bibitem{gayler_1998}
R.~Gayler, \emph{Multiplicative binding, representation operators \& analogy
  (workshop poster).} (New Bulgarian University, Sofia, 1998).
\newblock \url{http://cogprints.org/502/}

\bibitem{KanervaOrdered1996}
P.~Kanerva, \emph{{Binary Spatter-Coding of Ordered K-tuples}}, in
  \emph{{International Conference on Artificial Neural Networks (ICANN)}},
  \emph{Lecture Notes in Computer Science}, vol. 1112 (1996), pp. 869--873

\bibitem{eliasmith2003neural}
C.~Eliasmith, C.H. Anderson, \emph{Neural engineering: Computation,
  representation, and dynamics in neurobiological systems} (MIT press, 2003)

\bibitem{sandamirskaya2013dynamic}
Y.~Sandamirskaya, Dynamic neural fields as a step toward cognitive neuromorphic
  architectures.
\newblock Frontiers in Neuroscience \textbf{7}, 276 (2013).
\newblock \doi{10.3389/fnins.2013.00276}.

\bibitem{KleykoComputingParadigm2021}
D.~Kleyko, M.~Davies, E.P. Frady, et~al., {Vector Symbolic Architectures as a
  Computing Framework for Nanoscale Hardware}.
\newblock arXiv:2106.05268 (2021)

\bibitem{liang2019neural}
D.~Liang, R.~Kreiser, C.~Nielsen, N.~Qiao, Y.~Sandamirskaya, G.~Indiveri,
  Neural state machines for robust learning and control of neuromorphic agents.
\newblock IEEE Journal on Emerging and Selected Topics in Circuits and Systems
  \textbf{9}(4), 679--689 (2019)

\bibitem{GaylerJackendoff2003}
R.W. Gayler, \emph{{Vector Symbolic Architectures Answer Jackendoff's
  Challenges for Cognitive Neuroscience}}, in \emph{{Joint International
  Conference on Cognitive Science (ICCS/ASCS)}} (2003), pp. 133--138

\bibitem{kanerva_2009}
P.~Kanerva, Hyperdimensional computing: An introduction to computing in
  distributed representation with high-dimensional random vectors.
\newblock Cognitive computation \textbf{1}(2), 139--159 (2009).
\newblock \doi{10.1007/s12559-009-9009-8}.

\bibitem{plate1995holographic}
T.A. Plate, Holographic reduced representations.
\newblock {IEEE Transactions on Neural networks} \textbf{6}(3), 623--641 (1995)

\bibitem{frady_2019tpam}
E.P. Frady, F.T. Sommer, Robust computation with rhythmic spike patterns.
\newblock Proceedings of the National Academy of Sciences of the United States
  of America \textbf{116}(36), 18050--18059 (2019).
\newblock \doi{10.1073/pnas.1902653116}.

\bibitem{renner2022T}
A.~Renner, L.~Supic, A.~Danielescu, G.~Indiveri, B.A. Olshausen,
  Y.~Sandamirskaya, F.T. Sommer, E.P. Frady, Neuromorphic visual scene
  understanding with resonator networks.
\newblock Nature Machine Intelligence \textbf{6} (2024)
\newblock \doi{10.1038/s42256-024-00848-0}.
\newblock \url{https://arxiv.org/abs/2208.12880}

\bibitem{noest_1988}
A.~Noest, Phasor neural networks.
\newblock Neural information processing systems p. 584 (1988).
\newblock
  \url{http://papers.nips.cc/paper/90-phasor-neural-networks.pdf}

\bibitem{frady_2019precession}
P.~Frady, P.~Kanerva, F.~Sommer, A framework for linking computations and
  rhythm-based timing patterns in neural firing, such as phase precession in
  hippocampal place cells.
\newblock CCN  (2019)

\bibitem{frady2021VFA}
E.~Frady, D.~Kleyko, C.~Kymn, B.~Olshausen, F.~Sommer, Computing on functions
  using randomized vector representations.
\newblock {arXiv} preprint {arXiv}:2109.03429  (2021).
\newblock \url{https://arxiv.org/abs/2109.03429}

\bibitem{komer_2019}
B.~Komer, T.~Stewart, A.~Voelker, C.~Eliasmith, \emph{A neural representation
  of continuous space using fractional binding.}, in \emph{Annual Meeting of
  the Cognitive Science Society ({CogSci})} (Cognitive Science Society, 2019),
  pp. 2038--2043

\bibitem{frady_2020_resonator}
E.P. Frady, S.J. Kent, B.A. Olshausen, F.T. Sommer, Resonator networks, 1: An
  efficient solution for factoring high-dimensional, distributed
  representations of data structures.
\newblock Neural Computation pp. 1--21 (2020).
\newblock \doi{10.1162/neco\_a\_01331}.

\bibitem{kent_2020resonator}
S.J. Kent, E.P. Frady, F.T. Sommer, B.A. Olshausen, Resonator networks, 2:
  Factorization performance and capacity compared to optimization-based
  methods.
\newblock Neural Computation \textbf{32}(12), 2332--2388 (2020).
\newblock \doi{10.1162/neco\_a\_01329}.

\bibitem{casasent1976position}
D.~Casasent, D.~Psaltis, Position, rotation, and scale invariant optical
  correlation.
\newblock Applied optics \textbf{15}(7), 1795--1799 (1976)

\bibitem{chen1994symmetric}
Q.s. Chen, M.~Defrise, F.~Deconinck, Symmetric phase-only matched filtering of
  Fourier-Mellin transforms for image registration and recognition.
\newblock IEEE Transactions on pattern analysis and machine intelligence
  \textbf{16}(12), 1156--1168 (1994)

\bibitem{reddy1996fft}
B.S. Reddy, B.N. Chatterji, An FFT-based technique for translation, rotation,
  and scale-invariant image registration.
\newblock IEEE transactions on image processing \textbf{5}(8), 1266--1271
  (1996)

\bibitem{gallego2020event}
G.~Gallego, T.~Delbr{\"u}ck, G.~Orchard, C.~Bartolozzi, B.~Taba, A.~Censi,
  S.~Leutenegger, A.J. Davison, J.~Conradt, K.~Daniilidis, et~al., Event-based
  vision: A survey.
\newblock IEEE transactions on pattern analysis and machine intelligence
  \textbf{44}(1), 154--180 (2020)

\bibitem{reinbacher_2017}
C.~Reinbacher, G.~Munda, T.~Pock, \emph{Real-time panoramic tracking for event
  cameras}, in \emph{2017 {IEEE} International Conference on Computational
  Photography ({ICCP})} (IEEE, 2017), pp. 1--9.
\newblock \doi{10.1109/ICCPHOT.2017.7951488}.

\bibitem{rebecq2017real}
H.~Rebecq, T.~Horstschaefer, D.~Scaramuzza, \emph{Real-time Visual-Inertial
  Odometry for Event Cameras using Keyframe-based Nonlinear Optimization}, in
  \emph{Procedings of the British Machine Vision Conference 2017} (British
  Machine Vision Association, 2017), p.~16.
\newblock \doi{10.5244/C.31.16}.

\bibitem{zihao2017event}
A.~Zihao~Zhu, N.~Atanasov, K.~Daniilidis, \emph{Event-based visual inertial
  odometry}, in \emph{Proceedings of the IEEE Conference on Computer Vision and
  Pattern Recognition} (2017), pp. 5391--5399

\bibitem{vidal2018ultimate}
A.R. Vidal, H.~Rebecq, T.~Horstschaefer, D.~Scaramuzza, Ultimate slam?
  combining events, images, and imu for robust visual SLAM in HDR and
  high-speed scenarios.
\newblock IEEE Robotics and Automation Letters \textbf{3}(2), 994--1001 (2018)

\bibitem{nguyen_2019}
A.~Nguyen, T.T. Do, D.G. Caldwell, N.G. Tsagarakis, \emph{Real-Time {6DOF} Pose
  Relocalization for Event Cameras With Stacked Spatial {LSTM} Networks}, in
  \emph{2019 {IEEE}/{CVF} Conference on Computer Vision and Pattern Recognition
  Workshops ({CVPRW})} (IEEE, 2019), pp. 1638--1645.
\newblock \doi{10.1109/CVPRW.2019.00207}.

\bibitem{rebecq2019high}
H.~Rebecq, R.~Ranftl, V.~Koltun, D.~Scaramuzza, High speed and high dynamic
  range video with an event camera.
\newblock IEEE transactions on pattern analysis and machine intelligence
  (2019)

\bibitem{xiao2021research}
K.~Xiao, G.~Wang, Y.~Chen, Y.~Xie, H.~Li, Research on event accumulator
  settings for event-based slam.
\newblock arXiv preprint arXiv:2112.00427  (2021)

\bibitem{mueggler2017event}
E.~Mueggler, H.~Rebecq, G.~Gallego, T.~Delbruck, D.~Scaramuzza, The
  event-camera dataset and simulator: Event-based data for pose estimation,
  visual odometry, and slam.
\newblock The International Journal of Robotics Research \textbf{36}(2),
  142--149 (2017)

\bibitem{cook_2011}
M.~Cook, L.~Gugelmann, F.~Jug, C.~Krautz, A.~Steger, \emph{Interacting maps for
  fast visual interpretation}, in \emph{The 2011 International Joint Conference
  on Neural Networks} (IEEE, 2011), pp. 770--776.
\newblock \doi{10.1109/IJCNN.2011.6033299}.

\bibitem{kim_2014}
H.~Kim, A.~Handa, R.~Benosman, S.H. Ieng, A.~Davison, \emph{Simultaneous
  Mosaicing and Tracking with an Event Camera}, in \emph{Proceedings of the
  British Machine Vision Conference 2014} (British Machine Vision Association,
  2014), pp. 26.1--26.12.
\newblock \doi{10.5244/C.28.26}.

\bibitem{mueggler_2014}
E.~Mueggler, B.~Huber, D.~Scaramuzza, \emph{Event-based, 6-{DOF} pose tracking
  for high-speed maneuvers}, in \emph{2014 {IEEE}/{RSJ} International
  Conference on Intelligent Robots and Systems} (IEEE, 2014), pp. 2761--2768.
\newblock \doi{10.1109/IROS.2014.6942940}.

\bibitem{censi_2014}
A.~Censi, D.~Scaramuzza, \emph{Low-latency event-based visual odometry}, in
  \emph{2014 {IEEE} International Conference on Robotics and Automation
  ({ICRA})} (IEEE, 2014), pp. 703--710.
\newblock \doi{10.1109/ICRA.2014.6906931}.

\bibitem{weikersdorfer_2014}
D.~Weikersdorfer, D.B. Adrian, D.~Cremers, J.~Conradt, \emph{Event-based {3D}
  {SLAM} with a depth-augmented dynamic vision sensor}, in \emph{2014 {IEEE}
  International Conference on Robotics and Automation ({ICRA})} (IEEE, 2014),
  pp. 359--364.
\newblock \doi{10.1109/ICRA.2014.6906882}.

\bibitem{gallego_2015}
G.~Gallego, C.~Forster, E.~Mueggler, D.~Scaramuzza, Event-based camera pose
  tracking using a generative event model.
\newblock {arXiv} preprint {arXiv}:1510.01972  (2015).

\bibitem{kueng_2016}
B.~Kueng, E.~Mueggler, G.~Gallego, D.~Scaramuzza, \emph{Low-latency visual
  odometry using event-based feature tracks}, in \emph{2016 {IEEE}/{RSJ}
  International Conference on Intelligent Robots and Systems ({IROS})} (IEEE,
  2016), pp. 16--23.
\newblock \doi{10.1109/IROS.2016.7758089}.

\bibitem{rebecq_2017EMVS}
H.~Rebecq, G.~Gallego, E.~Mueggler, D.~Scaramuzza, {EMVS}: Event-based
  multi-view {Stereo\textemdash3D} reconstruction with an event camera in
  real-time.
\newblock International journal of computer vision \textbf{126}(12), 1--21
  (2017).
\newblock \doi{10.1007/s11263-017-1050-6}.

\bibitem{rebecq2016evo}
H.~Rebecq, T.~Horstsch{\"a}fer, G.~Gallego, D.~Scaramuzza, Evo: A geometric
  approach to event-based 6-DOF parallel tracking and mapping in real time.
\newblock IEEE Robotics and Automation Letters \textbf{2}(2), 593--600 (2016)

\bibitem{gallego2018unifying}
G.~Gallego, H.~Rebecq, D.~Scaramuzza, \emph{A unifying contrast maximization
  framework for event cameras, with applications to motion, depth, and optical
  flow estimation}, in \emph{Proceedings of the IEEE Conference on Computer
  Vision and Pattern Recognition} (2018), pp. 3867--3876

\bibitem{zhu_2019}
A.Z. Zhu, L.~Yuan, K.~Chaney, K.~Daniilidis, \emph{Unsupervised Event-Based
  Learning of Optical Flow, Depth, and Egomotion}, in \emph{2019 {IEEE}/{CVF}
  Conference on Computer Vision and Pattern Recognition ({CVPR})} (IEEE, 2019),
  pp. 989--997.
\newblock \doi{10.1109/CVPR.2019.00108}.

\bibitem{davies2018loihi}
M.~Davies, N.~Srinivasa, T.H. Lin, G.~Chinya, Y.~Cao, S.H. Choday, G.~Dimou,
  P.~Joshi, N.~Imam, S.~Jain, et~al., Loihi: A neuromorphic manycore processor
  with on-chip learning.
\newblock IEEE Micro \textbf{38}(1), 82--99 (2018)

\bibitem{EsserEtAl2016}
S.~Esser, P.~Merolla, J.~Arthur, A.~Cassidy, R.~Appuswamy, A.~Andreopoulos,
  D.~Berg, J.~McKinstry, T.~Melano, D.~Barch, C.~di~Nolfo, D.~P., A.~Amir,
  B.~Taba, M.~Flickner, D.~Modha, Convolutional networks for fast,
  energy-efficient neuromorphic computing.
\newblock PNAS \textbf{113}, 11441--11446 (2016)

\bibitem{frenkel20180}
C.~Frenkel, M.~Lefebvre, J.D. Legat, D.~Bol, A 0.086-mm $^2 $12.7-pj/sop
  64k-synapse 256-neuron online-learning digital spiking neuromorphic processor
  in 28-nm cmos.
\newblock IEEE transactions on biomedical circuits and systems \textbf{13}(1),
  145--158 (2018)

\bibitem{shrestha2018slayer}
S.B. Shrestha, G.~Orchard, \emph{Slayer: Spike layer error reassignment in
  time}, in \emph{Advances in Neural Information Processing Systems} (2018),
  pp. 1412--1421

\bibitem{renner2021backpropagation}
A.~Renner, F.~Sheldon, A.~Zlotnik, L.~Tao, A.~Sornborger, The backpropagation
  algorithm implemented on spiking neuromorphic hardware.
\newblock arXiv preprint arXiv:2106.07030  (2021)

\bibitem{frady_2020knn}
E.P. Frady, G.~Orchard, D.~Florey, N.~Imam, R.~Liu, J.~Mishra, J.~Tse, A.~Wild,
  F.T. Sommer, M.~Davies, \emph{Neuromorphic nearest neighbor search using
  intel's pohoiki springs}, in \emph{Proceedings of the Neuro-inspired
  Computational Elements Workshop} (ACM, New York, {NY}, {USA}, 2020), pp.
  1--10.
\newblock \doi{10.1145/3381755.3398695}.

\bibitem{serranogotarredona_2009}
R.~Serrano-Gotarredona, M.~Oster, P.~Lichtsteiner, A.~Linares-Barranco,
  R.~Paz-Vicente, F.~Gomez-Rodriguez, L.~Camunas-Mesa, R.~Berner,
  M.~Rivas-Perez, T.~Delbruck, S.C. Liu, R.~Douglas, P.~Hafliger,
  G.~Jimenez-Moreno, A.~Civit~Ballcels, T.~Serrano-Gotarredona, A.J.
  Acosta-Jimenez, B.~Linares-Barranco, {CAVIAR}: a 45k neuron, {5M} synapse,
  {12G} connects/s {AER} hardware sensory-processing- learning-actuating system
  for high-speed visual object recognition and tracking.
\newblock {IEEE} Transactions on Neural Networks \textbf{20}(9), 1417--1438
  (2009).
\newblock \doi{10.1109/TNN.2009.2023653}.

\bibitem{kendall_2015}
A.~Kendall, M.~Grimes, R.~Cipolla, \emph{{PoseNet}: A Convolutional Network for
  Real-Time 6-{DOF} Camera Relocalization}, in \emph{2015 {IEEE} International
  Conference on Computer Vision ({ICCV})} (IEEE, 2015), pp. 2938--2946.
\newblock \doi{10.1109/ICCV.2015.336}.

\bibitem{kendall_2016}
A.~Kendall, R.~Cipolla, \emph{Modelling uncertainty in deep learning for camera
  relocalization}, in \emph{2016 {IEEE} International Conference on Robotics
  and Automation ({ICRA})} (IEEE, 2016), pp. 4762--4769.
\newblock \doi{10.1109/ICRA.2016.7487679}.

\bibitem{walch_2017}
F.~Walch, C.~Hazirbas, L.~Leal-Taixe, T.~Sattler, S.~Hilsenbeck, D.~Cremers,
  \emph{Image-Based Localization Using {LSTMs} for Structured Feature
  Correlation}, in \emph{2017 {IEEE} International Conference on Computer
  Vision ({ICCV})} (IEEE, 2017), pp. 627--637.
\newblock \doi{10.1109/ICCV.2017.75}.

\bibitem{laskar_2017}
Z.~Laskar, I.~Melekhov, S.~Kalia, J.~Kannala, \emph{Camera relocalization by
  computing pairwise relative poses using convolutional neural network}, in
  \emph{2017 {IEEE} International Conference on Computer Vision Workshops
  ({ICCVW})} (IEEE, 2017), pp. 920--929.
\newblock \doi{10.1109/ICCVW.2017.113}.

\bibitem{arathorn2002map}
D.W. Arathorn, \emph{Map-seeking circuits in visual cognition: A computational
  mechanism for biological and machine vision} (Stanford University Press,
  2002)

\bibitem{vogel2006retinal}
C.R. Vogel, D.W. Arathorn, A.~Roorda, A.~Parker, Retinal motion estimation in
  adaptive optics scanning laser ophthalmoscopy.
\newblock Optics express \textbf{14}(2), 487--497 (2006)

\bibitem{bowman2017probabilistic}
S.L. Bowman, N.~Atanasov, K.~Daniilidis, G.J. Pappas, \emph{Probabilistic data
  association for semantic slam}, in \emph{2017 IEEE international conference
  on robotics and automation (ICRA)} (IEEE, 2017), pp. 1722--1729

\bibitem{brandli_2014}
C.~Brandli, R.~Berner, M.~Yang, S.C. Liu, T.~Delbruck, A 240 x 180 130 {dB} 3
  us latency global shutter spatiotemporal vision sensor.
\newblock {IEEE} Journal of Solid-State Circuits \textbf{49}(10), 2333--2341
  (2014).
\newblock \doi{10.1109/JSSC.2014.2342715}.

\bibitem{plate_1994}
T.~Plate, \emph{Distributed representations and nested compositional structure}
  (University of Toronto, Department of Computer Science, 1994).
\newblock
\url{http://citeseerx.ist.psu.edu/viewdoc/download?doi=10.1.1.48.5527\&rep=rep1\&type=pdf}

\bibitem{mitrokhin_2019}
A.~Mitrokhin, P.~Sutor, C.~Fermüller, Y.~Aloimonos, Learning sensorimotor
  control with neuromorphic sensors: Toward hyperdimensional active perception.
\newblock Science Robotics \textbf{4}(30) (2019).
\newblock \doi{10.1126/scirobotics.aaw6736}.

\bibitem{renner_2019}
A.~Renner, M.~Evanusa, Y.~Sandamirskaya, \emph{Event-Based Attention and
  Tracking on Neuromorphic Hardware}, in \emph{2019 {IEEE}/{CVF} Conference on
  Computer Vision and Pattern Recognition Workshops ({CVPRW})} (IEEE, 2019),
  pp. 1709--1716.
\newblock \doi{10.1109/CVPRW.2019.00220}.

\bibitem{umeyama1991least}
S.~Umeyama, Least-squares estimation of transformation parameters between two
  point patterns.
\newblock IEEE Transactions on Pattern Analysis \& Machine Intelligence
  \textbf{13}(04), 376--380 (1991)

\bibitem{Zhang18iros}
Z.~Zhang, D.~Scaramuzza, \emph{A Tutorial on Quantitative Trajectory Evaluation
  for Visual(-Inertial) Odometry}, in \emph{IEEE/RSJ Int. Conf. Intell. Robot.
  Syst. (IROS)} (2018)

\bibitem{renner_2024_codeocean_vo}
A.~Renner, L.~Supic, E.P. Frady.
\newblock Code for visual odometry with neuromorphic resonator networks.
\newblock Codeocean (2024).
\newblock \doi{10.24433/CO.6568112.v1}.

\end{thebibliography}
\end{document}